\documentclass[10pt,twocolumn,letterpaper]{article}

\usepackage{cvpr}              % To produce the CAMERA-READY version
\usepackage{graphicx}
\usepackage{amsmath}
\usepackage{amssymb}
\usepackage{booktabs}

\usepackage[pagebackref,breaklinks,colorlinks]{hyperref}

\usepackage[capitalize]{cleveref}
\crefname{section}{Sec.}{Secs.}
\Crefname{section}{Section}{Sections}
\Crefname{table}{Table}{Tables}
\crefname{table}{Tab.}{Tabs.}

\def\cvprPaperID{5389} % *** Enter the CVPR Paper ID here
\def\confName{CVPR}
\def\confYear{2023}

\usepackage{amsmath,amsfonts,bm}

\def\eqref#1{equation~\ref{#1}}
\def\1{\bm{1}}

\DeclareMathAlphabet{\mathsfit}{\encodingdefault}{\sfdefault}{m}{sl}
\SetMathAlphabet{\mathsfit}{bold}{\encodingdefault}{\sfdefault}{bx}{n}

\usepackage[utf8]{inputenc} % allow utf-8 input
\usepackage[T1]{fontenc}    % use 8-bit T1 fonts
\usepackage{hyperref}       % hyperlinks
\hypersetup{breaklinks=true,colorlinks,citecolor=blue}
\usepackage{url}            % simple URL typesetting
\usepackage{booktabs}       % professional-quality tables
\usepackage{amsfonts}       % blackboard math symbols
\usepackage{nicefrac}       % compact symbols for 1/2, etc.
\usepackage{microtype}      % microtypography
\usepackage[table,xcdraw]{xcolor}
\usepackage{subcaption}
\usepackage{comment}
\usepackage{graphicx}
\usepackage{floatrow}
\usepackage{bm}
\usepackage{wrapfig}
\usepackage{amsmath}
\usepackage{multirow}
\usepackage{graphicx,paralist,booktabs,tabu}
 
\newcommand{\topic}[1]{\vspace{1mm}\noindent\textbf{#1}}

\usepackage{makecell}

\usepackage{amsthm}
\newtheorem{theorem}[]{Theorem}

\title{Hyperbolic Contrastive Learning for Visual Representations beyond Objects}

\newcommand\blfootnote[1]{%
  \begingroup
  \renewcommand\thefootnote{}\footnote{#1}%
  \addtocounter{footnote}{-1}%
  \endgroup
}

\begin{document}

\author{Songwei Ge$^*$\textsuperscript{\rm 1},
% University of Maryland College Park\\
% {\tt\small shlokm@cs.umd.edu}
% \and
Shlok Mishra$^*$\textsuperscript{\rm 1},
Simon Kornblith\textsuperscript{\rm 2},\\
Chun-Liang Li\textsuperscript{\rm 2},
David Jacobs\textsuperscript{\rm 1,3} 
\\
\textsuperscript{\rm 1}University of Maryland, College Park,
\textsuperscript{\rm 2}Google Research,\\
    % \textsuperscript{\rm 3}Google Research, \\
    \textsuperscript{\rm 3}Meta
\texttt{\{songweig,shlokm,dwj\}@umd.edu} 
}

\maketitle

\begin{abstract}
 Although self-/un-supervised methods have led to rapid progress in visual representation learning, these methods generally treat objects and scenes using the same lens. In this paper, we focus on learning representations for objects and scenes that preserve the structure among them.
 Motivated by the observation that visually similar objects are close in the representation space, we argue that the scenes and objects should instead follow a hierarchical structure based on their compositionality. To exploit such a structure, we propose a contrastive learning framework where a Euclidean loss is used to learn object representations and a hyperbolic loss is used to encourage representations of scenes to lie close to representations of their constituent objects in a hyperbolic space. This novel hyperbolic objective encourages the scene-object hypernymy among the representations by optimizing the magnitude of their norms. We show that when pretraining on the COCO and OpenImages datasets, the hyperbolic loss improves downstream performance of several baselines across multiple datasets and tasks, including image classification, object detection, and semantic segmentation. We also show that the properties of the learned representations allow us to solve various vision tasks that involve the interaction between scenes and objects in a zero-shot fashion. Our code can be found at \url{https://github.com/shlokk/HCL/tree/main/HCL}.
\end{abstract}
\blfootnote{$^*$Equal Contribution. The order is decided randomly.}

\section{Introduction}

% \dwj{Overall, this looks good.  I think the main issues you need to address are providing a more intuitive description of hyperbolic space and how you use it, and making clear what the novel contribution is.}

Our visual world is diverse and structured. Imagine taking a close-up of a box of cereal in the morning.  If we zoom out slightly, we may see different nearby objects such as a pitcher of milk, a cup of hot coffee, today's newspaper, or reading glasses. Zooming out further, we will probably recognize that these items are placed on a dining table with the kitchen as background rather than inside a bathroom. Such scene-object structure is diverse, yet not completely random. In this paper, we aim at learning visual representations of both the cereal box (objects) and the entire dining table (scenes) in the same space while preserving such hierarchical structures.

% \begin{wrapfigure}[23]{r}{0.47\textwidth}
% \centering
\begin{figure}[t]
\begin{center}
  \includegraphics[width=\textwidth]{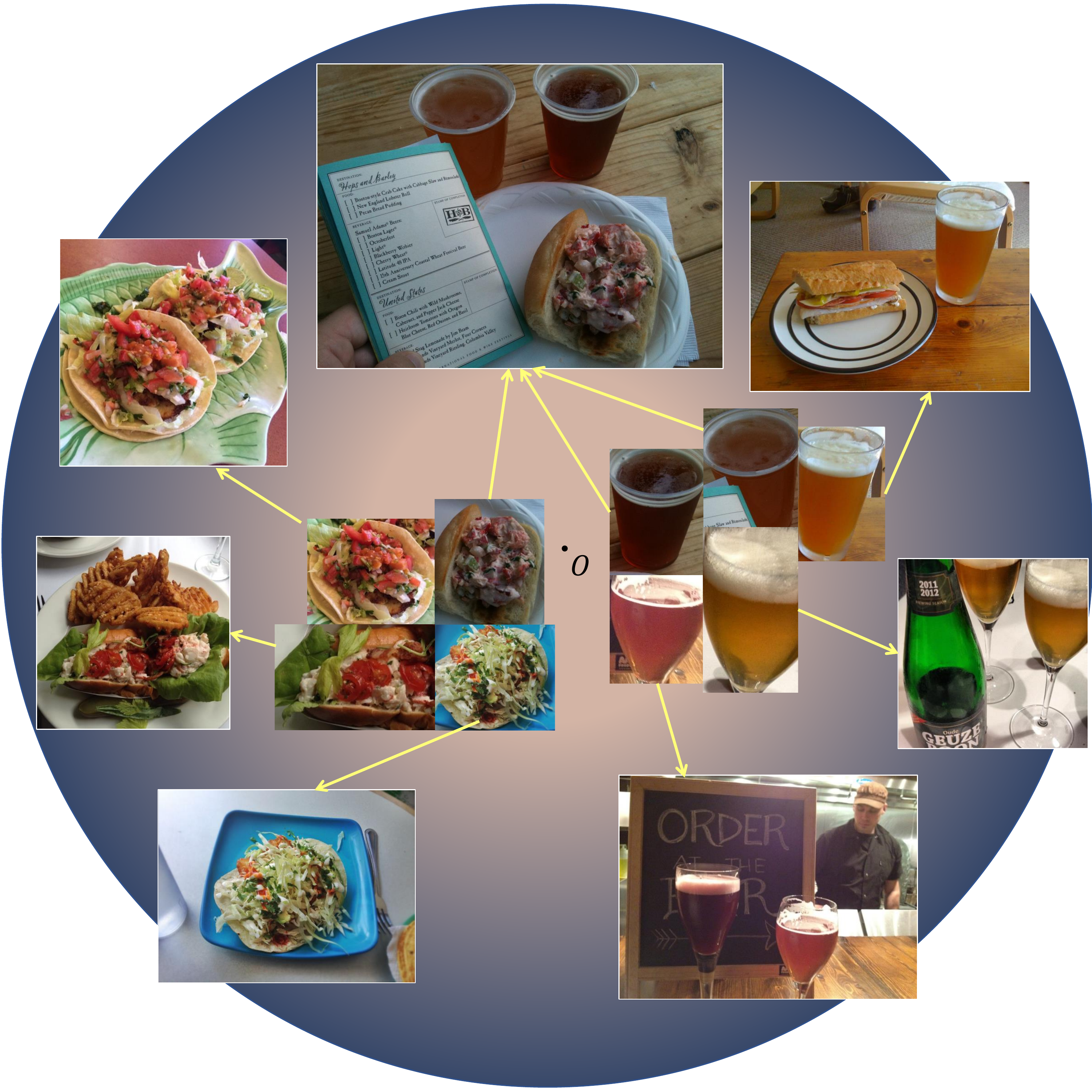}
\end{center}  
  \captionof{figure}{Illustration of the representation space learned by our models. Object images of the same class tend to gather near the center around similar directions, while the scene images are far away in these directions with larger norms.}
  \label{fig:teaser}
  
\end{figure}

Un-/self-supervised learning has become a standard method to learn visual representations~\cite{he2020momentum,chen2020simple,grill2020bootstrap,caron2021emerging,radford2021learning,He_2022_CVPR}. Although these methods attain superior performance over supervised pretraining on object-centric datasets such as ImageNet~\cite{caron2020unsupervised}, inferior results are observed on images depicting multiple objects such as OpenImages or COCO~\cite{xie2021unsupervised}. Several methods have been proposed to mitigate this issue, but all focus either on learning improved object representations~\cite{xie2021unsupervised,bai2022point} or dense pixel representations~\cite{xie2021propagate,liu2021selfemd,wang2021dense}, instead of explicitly modeling representations for scene images. The object representations learned by these methods present a natural topology~\cite{wu2018unsupervised}. That is, the objects from visually similar classes lie close to each other in the representation space. However, it is not clear how the representations of scene images should fit into that topology. Directly applying existing contrastive learning results in a sub-optimal topology of scenes and objects as well as unsatisfactory performance, as we will show in the experiments.  To this end, we argue that a hierarchical structure can be naturally adopted. 
% Considering scenes as the composition of different kinds of objects,
Considering that the same class of objects can be placed in different scenes,
we construct a hierarchical structure to describe such relationships, where the root nodes are the visually similar objects, and the scene images consisting of them are placed as the descendants. We call this structure the object-centric scene hierarchy.

The intermediate modeling difficulty induced by this structure is the combinatorial explosion. A finite number of objects leads to exponentially many different possible scenes. Consequently, Euclidean space may require an arbitrarily large number of dimensions to faithfully embed these scenes, whereas it is known that any infinite trees can be embedded without distortion in a 2D hyperbolic space~\cite{gromov1987hyperbolic}.
% Hyperbolic space is known for its provably better capacity in modeling infinite trees with significantly lower distortion compared with Euclidean space~\cite{ganea2018hyperbolic,gromov1987hyperbolic,linial1995geometry}. 
Therefore, we propose to employ a hyperbolic objective to regularize the scene representations.
%Our framework is based on a contrastive learning model~\cite{he2020momentum,chen2020simple}, which is built to learn object representations.
% Our framework builds upon MoCo~\cite{he2020momentum}, which has been shown to learn good object representations. 
To learn representations of scenes, in the general setting of contrastive learning, we sample co-occurring scene-object pairs as positive pairs, and objects that are not part of that scene as negative samples, and use these pairs to compute an auxiliary hyperbolic contrastive objective. Our model is trained to reduce the distance between positive pairs and push away the negative pairs in a hyperbolic space.

Contrastive learning usually has objectives defined on a hypersphere~\cite{he2020momentum,chen2020simple}. By discarding the norm information, these models circumvent the shortcut of minimizing losses through tuning the norms and obtain better downstream performance.
%At the same time, they lose freedom of the representative power in the magnitude of the norm and leave the images disorganized.
However, the norm of the representation can also be used to encode useful representational structure.
In hyperbolic space, the magnitude of a vector often plays the role of modeling the hypernymy of the hierarchical structure~\cite{nickel2017poincare,tifrea2018poincare,sala2018representation}. When projecting the representations to the hyperbolic space, the norm information is preserved and used to determine the Riemannian distance, which eventually affects the loss. Since hyperbolic space is diffeomorphic and conformal to Euclidean space, our hyperbolic contrastive loss is  differentiable and complementary to the original contrastive objective.

When training simultaneously with the original contrastive objective for objects and our proposed hyperbolic contrastive objective for scenes, the resulting representation space exhibits a desired hierarchical structure while leaving the object clustering topology intact as shown in Figure~\ref{fig:teaser}. We demonstrate the effectiveness of the 
% learned representations 
hyperbolic objective under several frameworks on multiple downstream tasks. We also show that the properties of the representations allow us to perform various vision tasks in a zero-shot way, from label uncertainty quantification to out-of-context object detection. Our contributions are summarized below:

\begin{enumerate}
    % \item We propose to learn representations for both object and scene images simultaneously using un-/self-supervised methods.
    % % . \dwj{This is a pretty strong statement.  Are there really *no* such works?} 
    % % \shlok{probably should say using hyperoblic loss}
    % We propose to explore an object-centric scene hierarchy that the representations are expected to follow.
    % % \item We propose a framework with a novel hyperbolic contrastive loss to regularize the scene representations with positive and negative pairs sampled from the hierarchy.
    % \item We propose a novel hyperbolic contrastive loss to regularize the scene representations with positive and negative pairs sampled from the hierarchy.
    \item We propose a hyperbolic contrastive loss that regularizes scene  representations so that they follow an object-centric hierarchy, with positive and negative pairs sampled from the hierarchy.
    \item We demonstrate that our learned representations transfer better than representations learned using vanilla contrastive loss on a variety of downstream tasks, including object detection, semantic segmentation, and linear classification.
    \item We show that the magnitude of representation norms effectively reflect the scene-objective hypernymy.
\end{enumerate}

\section{Method}
In this section, we elaborate upon our approach to learning visual representations of object and scene images. We start by describing the hierarchical structure between objects and scenes that we wish to enforce in the learned representation space.

\subsection{Object-Centric Scene Hierarchy}
From simple object co-occurrence statistics~\cite{galleguillos2008object,mensink2014costa} to finer object relationships~\cite{johnson2015image,krishna2017visual}, using hierarchical relationships between objects and scenes to understand images is not new. Previous studies primarily work on an image-level hierarchy by dividing an image into its lower-level elements recursively: a scene contains multiple objects, an object has different parts, and each part may consist of even lower-level features~\cite{parikh2007hierarchical,choi2010exploiting,hinton2021represent}. While this is intuitive, it describes a hierarchical structure contained in the individual images. 
% In our task, we would like to learn a single representation space that is shared by all objects and scenes across the entire dataset. To this end, we argue that it is more natural to consider an object-centric hierarchy.
Instead, we study the structure presented among different images. Our goal is to learn a representation space for images of both objects and scenes across the entire dataset. To this end, we argue that it is more natural to consider an \textit{object-centric hierarchy}.

It is known that when training an image classifier, the objects from visually similar classes often lie close to each other in the representation space~\cite{wu2018unsupervised}, which has become the cornerstone of contrastive learning.
Motivated by this observation, we believe that the representation of each scene image should also be close to the object clusters it consists of. However, modeling scenes requires a much larger volume due to the exponential number of possible compositions of objects. Another way to think about the object-centric hierarchy is through the generality and specificity as often discussed in the language literature~\cite{miller1990introduction,nickel2017poincare}. An object concept is general when standing alone in the visual world, and it will become specific when a certain context is given. For example, ``a desk'' is thought to be a more general concept than ``a desk in a classroom with a boy sitting on it''. 

Therefore, we propose to study an object-centric hierarchy across the entire dataset. Formally, given a set of images $\mathcal{S} = \{s_1, s_2, \cdots, s_n\}$, $\mathcal{O}_i = \{o_i^1, o_i^2, \cdots, o_i^{n_i}\}$ are the object bounding boxes contained in the image $s_i$. We define the regions of scene $\mathcal{R}_i = \{r_i^1, r_i^2, \cdots, r_i^{m_i}\}$ to be partial areas of the image $s_i$ that contain multiple objects such that $r_i^j = \displaystyle\cup_k  o_i^k$, where $o_i^k\in \mathcal{O}_i \text{ and object $k$ is in the region } j$. We define the object-centric hierarchy $T = (V, E)$ to be that $V = \mathcal{S} \displaystyle\cup \mathcal{O} \displaystyle\cup \mathcal{R}$, where $ \mathcal{R} =\mathcal{R}_1\cup \cdots \cup \mathcal{R}_n$ and $\mathcal{O} =\mathcal{O}_1\cup \cdots \cup \mathcal{O}_n$. For $u, v\in V$, $e=(u, v)$ is an edge of $T$ if $u \subseteq v$ or $v \subseteq u$. Note that the natural scene images $\mathcal{S}$ are always put as the leaf nodes.

\subsection{Representation Learning beyond Objects}
To describe our proposed model based on this hierarchy, we begin with a brief review of hyperbolic space and its properties used in our model. For comprehensive introductions to Riemannian geometry and hyperbolic space, we refer the readers to \cite{do1992riemannian,lee2018introduction}. 
\subsubsection{Hyperbolic Space}

A hyperbolic space $(\mathbb{H}^m, g)$ is a complete, connected Riemannian manifold with constant negative sectional curvature. These special manifolds are all isometric to each other with the isometries defined as $O^+(m, 1)$. Among these isometries, there are five common models that previous studies often work on~\cite{cannon1997hyperbolic}. In this paper, we choose the Poincaré ball $\mathbb{D}^{n}:=\left\{p \in \mathbb{R}^{n} \mid \|p\|^{2}<r^2\right\}$ as our basic model~\cite{nickel2017poincare,tifrea2018poincare,ganea2018hyperbolicn}, where $r>0$ is the radius of the ball. The Poincaré ball is coupled with a Riemannian metric $g_\mathbb{D}(p)=\frac{4}{\left(1-\|p\|^{2}/r^2\right)^{2}} g_\mathbb{E}$, where $p \in \mathbb{D}^n$ and $g_\mathbb{E}$ is the canonical metric of the Euclidean space. For $p, q \in \mathbb{D}$, the Riemannian distance on the Poincaré ball induced by its metric $g_\mathbb{D}$ is defined as follows:
\begin{equation}
\label{eq:hyp_dis}
    d_\mathbb{D}(p, q)=2r \tanh ^{-1}\left(\frac{\left\|- p \oplus q\right\|}{r}\right),
\end{equation}
where $\oplus$ is the Möbius addition and it is clearly differentiable. In addition, the Poincaré ball can be viewed as a natural counterpart of the hypersphere as it allows all directions, unlike the other models such as the halfspace or hemisphere models that have constraints on the directions. The hyperbolic space is globally differomorphic to the Euclidean space, which is stated in the theorem below:

\begin{theorem}
\textbf{(Cartan–Hadamard).}  For every point $p \in \mathbb{H}^n$ the exponential map $\exp _{p}: T_{p} \mathbb{H}^n \approx \mathbb{R}^n\rightarrow \mathbb{H}^n$ is a smooth covering map. Since $\mathbb{H}^n$ is simply connected, it is diffeomorphic to $\mathbb{R}^n$.
\end{theorem}

Specifically, for $p \in \mathbb{D}^n$ and $v \in T_{p} \mathbb{D}^n \approx \mathbb{R}^n$, the exponential map of the Poincaré ball  $\exp_{p}: T_{p} \mathbb{D}^n \rightarrow \mathbb{D}^n$ is defined as 
\begin{equation}
\label{eq:exp_map}
    \exp _{p}(v):=p \oplus\left(\tanh \left(\frac{r\|v\|}{r^2-\|p\|^{2}}\right) \frac{rv}{\|v\|}\right),
\end{equation} 
The exponential map gives us a way to map the output of a network, which is in the Euclidean space, to the Poincaré ball. In practice, to avoid numerical issues, we clip the maximal norm of $v$ with $r-\varepsilon$ before the projection, where $\varepsilon>0$. During the backpropagation, we perform RSGD~\cite{bonnabel2013stochastic} by scaling the gradients by $g_\mathbb{D}(p)^{-1}$. Intuitively, this forces the optimizer to take a smaller step when $p$ is closer to the boundary. The scaling factor is lower bounded by $\mathcal{O}(\varepsilon^2)$.

The immediate consequence of the negative curvature is that for any point $\boldsymbol{p}\in\mathbb{H}^m$, there are no conjugate points along any geodesic starting from $\boldsymbol{p}$. Therefore, the volume grows exponentially faster in hyperbolic space than in Euclidean space. Such a property makes it suitable to embed the hierarchical structure that has constant branching factors and exponential number of nodes. This is formally stated in the theorem below:

\begin{theorem}\cite{gromov1987hyperbolic}
Given a Poincaré ball $\mathbb{D}^n$ with an arbitrary dimension $n \ge 2$ and any set of points $p_1,\cdots, p_m \in \mathbb{D}^n$, there exists a
finite weighted tree $(T, d_{T})$ and an embedding $f: T \rightarrow \mathbb{D}^n$
such that for all i, j,
$$\left|d_{T}\left(f^{-1}\left(x_{i}\right), f^{-1}\left(x_{j}\right)\right)-d_\mathbb{D}\left(x_{i}, x_{j}\right)\right|=\mathcal{O}(\log (1+\sqrt{2}) \log (m))$$
\end{theorem}

Intuitively, the theorem states that any tree can be embedded into a Poincaré disk ($n=2$) with low distortion. On the contrary, it is known that the Euclidean space with unbounded number of dimensions is not able to achieve such a low distortion~\cite{linial1995geometry}. One useful intuition~\cite{sala2018representation} to help understand the advantage of the hyperbolic space is given two points $p, q \in \mathbb{D}^n$ s.t. $\|p\|=\|q\|$,  
\begin{equation}
    \label{eq:intuition}
    d_\mathbb{D}(p, q) \rightarrow d_\mathbb{D}(p, 0) + d_\mathbb{D}(0, q), \text{ as } \|p\|=\|q\| \rightarrow r
\end{equation}
This property basically reflects the fact that the shortest path in a tree is the path through the earliest common ancestor, and it is reproduced in the Poincaré when points are both close to the boundary.

% \dwj{This seems to mean that two scenes, no matter how similar, will be far apart?}
% \sg{Not necessary - this only holds in the extreme case that points are very close to the boundary. And when scenes are slightly away from the boundary and have small angles between them, the distance could still be very small.}

% \dwj{I think this section is basically good, but is lacking an intuitive summary.  Is there any way to give more intuitions about this space? }

\subsubsection{Hyperbolic Contrastive Learning}

\begin{figure*}[t]
    \centering
    \includegraphics[width=0.9\linewidth]{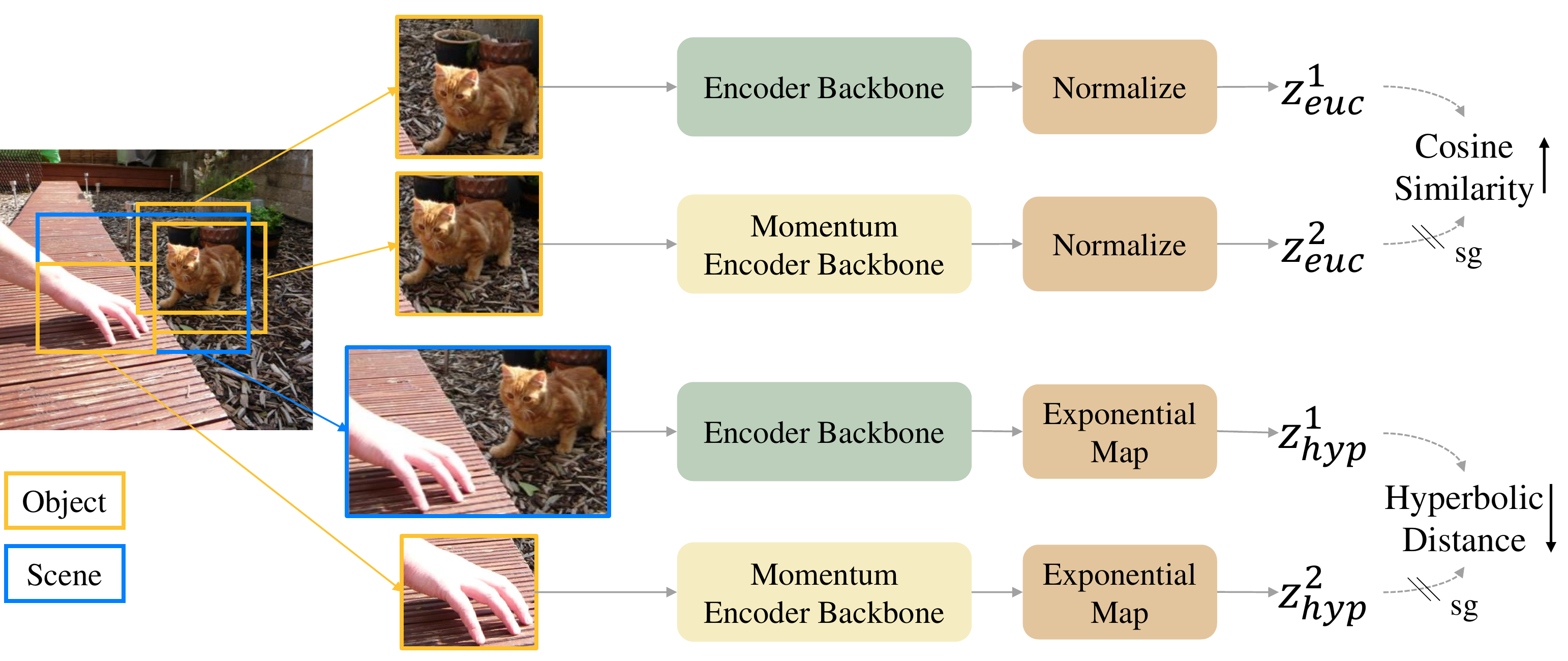}
    \caption{Our \textbf{H}yperbolic \textbf{C}ontrastive \textbf{L}earning (HCL) framework has two branches: given a scene image, two object regions are cropped to learn the object representations with a loss defined in the Euclidean space focusing on the representation directions. A scene region as well as a contained object region are used to learn the scene representations with a loss defined in the hyperbolic space that affects the representation norms.}
    \label{fig:model}
\end{figure*}

Given the theoretical benefits of the hyperbolic space stated above, we propose a contrastive learning framework as shown in Figure~\ref{fig:model}. We adopt two losses to learn the object and scene representations. First, to learn object representations, we use the standard normalized temperature-scaled cross-entropy loss, which operates on the hypersphere in Euclidean space. As shown in the top branch of Figure~\ref{fig:model}, we crop two views of a jittered and slightly expanded object region as the positive pairs and feed into the base and momentum encoders to calculate the object representations. We denote the output after the normalization to be $\mathbf{z}_{\text{euc}}^1$ and $\mathbf{z}_{\text{euc}}^2$. 
% Given the computational cost of large batch sizes, 
We follow MoCo~\cite{he2020momentum} and leverage a memory bank to store the negative representations $\boldsymbol{z}^{n}_{\text{euc}}$, which are the features $\mathbf{z}_{\text{euc}}^2$ from the previous batches. Note that our framework can be readily extended to other contrastive learning models. The Euclidean loss for each image is then calculated as:
\begin{equation*}
    \mathcal{L}_{\text{euc}} = -\log \frac{\exp \left(\mathbf{z}_{\text{euc}}^1\cdot \mathbf{z}_{\text{euc}}^2 / \tau\right)}{\exp \left(\mathbf{z}_{\text{euc}}^1\cdot \mathbf{z}_{\text{euc}}^2 / \tau\right)+\sum_{n} \exp \left(\mathbf{z}_{\text{euc}}^1\cdot \mathbf{z}_{\text{euc}}^{n} / \tau\right)},
\end{equation*}
where $\tau$ is a temperature parameter. 

While the loss above aims to learn object representations, we propose a hyperbolic contrastive objective to learn the representations for scene images. We sample positive region pairs $u$ and $v$ from object-centric scene hierarchy $T$ such that $(u, v) \in E$. In other words, as shown in the bottom branch of Figure~\ref{fig:model}, the objects contained in one region are required to be a subset of the objects in the other. We sample the negative samples of $u$ to be $\mathcal{N}_u = \{v | (u, v) \not\in E \}$. However, building and sampling exhaustively from the entire hierarchy explicitly is tricky. 
% Instead, we assume that there are exponentially more scenes than object classes.
In practice, given an image $s$, we always sample $u\in \mathcal{R} \cup \{s\}$ to be a scene region, $v\in \mathcal{O}$ to be an object that occurs in $u$, and $\mathcal{N}_u$ to be the other objects that are not in $u$.

The pair of scene and object images are fed into the base and momentum encoders that share the weights with the Euclidean branch. However, instead of normalizing the output of the encoders, we use the exponential map defined in the~\eqref{eq:exp_map} to project these features in the Euclidean space to the Poincaré ball, which are denoted as $\mathbf{z}_{\text{hyp}}^1$ and $\mathbf{z}_{\text{hyp}}^2$. Further, we replace the inner product in the cross-entropy loss with the negative hyperbolic distance as defined in~\eqref{eq:hyp_dis}. We calculate the hyperbolic contrastive loss as follows:
\begin{equation*}
    \mathcal{L}_{\text{hyp}} = -\log \frac{\exp \left(-\frac{d_\mathbb{D}(\mathbf{z}_{\text{hyp}}^1, \mathbf{z}_{\text{hyp}}^2)}{\tau}\right)}{\exp \left(- \frac{d_\mathbb{D}(\mathbf{z}_{\text{hyp}}^1, \mathbf{z}_{\text{hyp}}^2)}{\tau}\right)+\sum_{n} \exp \left(- \frac{d_\mathbb{D}(\mathbf{z}_{\text{hyp}}^1, \mathbf{z}_{\text{hyp}}^{n})}{\tau}\right)},
\end{equation*}
When minimizing the distances of all the positive pairs, with the intuition from~\eqref{eq:intuition}, it would be beneficial to put the nodes near the root, i.e. objects, close to the center to achieve an overall lower loss.
The overall loss function of our model is as follows:
\begin{equation*}
    \mathcal{L} = \mathcal{L}_{\text{euc}} + \lambda\mathcal{L}_{\text{hyp}},
\end{equation*}
where $\lambda$ is a scaling parameter to control the trade-off between hyperbolic and Euclidean losses.

\section{Experiments}

\subsection{Implementation Details}

\topic{Pre-training phase.} We pre-train on three datasets: COCO~\cite{Lin2014MicrosoftCC}, the full OpenImages labelled dataset \cite{kuznetsova2020open}($\sim 1.7$ million samples) and a subset of OpenImages ($\sim 212k$) \cite{Mishra2021ObjectAwareCF}. All these datasets are multi-object datasets; OpenImages 
% ($\sim212$k images) 
contains 12 objects on average per image and COCO 
% ($\sim118$k) 
contains 6 objects on average. We experiment with both the ground truth bounding box (GT) and using selective search (SS)~\cite{uijlings2013selective} to produce object bounding boxes in an unsupervised fashion, following previous work~\cite{xie2021unsupervised}. As the goal of this paper is not to present another state-of-the-art self-supervised learning method, we implement our sampling procedure and hyperbolic loss on top of three popular contrastive learning methods: MoCo-v2~\cite{chen2020improved}, Dense-CL~\cite{wang2021dense}, and ORL~\cite{xie2021unsupervised}. Dense-CL is a contrastive learning framework which extracts dense features from scene images and generally achieves better object detection results than MoCo-v2. ORL is a pipeline that learns improved object representations from scene images. We also consider HCL without the hyperbolic loss $\mathcal{L}_{\text{hyp}}$. This approach, which we denote as ``HCL w/o $\mathcal{L}_{\text{hyp}}$'', adopts the same cropping strategy as HCL but applies only a standard contrastive loss. We show that adding the hyperbolic loss improves results under various settings. More details on the datasets as well as training setups can be found in Appendix A. 

\label{sec:dataset}
\topic{Downstream tasks.} We evaluate our pre-trained models on image classification, object-detection and semantic segmentation. For classification, we show linear evaluation (lineval) accuracy with MoCo-v2, i.e. we freeze the backbone and only train the final linear layer. We test on VOC~\cite{everingham2010pascal}, ImageNet-100~\cite{tian2019contrastive} and ImageNet-1k~\cite{imagenet_cvpr09} datasets. 
% To test the discriminative capacity of the representations on both objects and scenes, we create a dataset by mixing the ImageNet-100 and a subset of Place-205~\cite{zhou2014learning} datasets, which we refer to as the INPMix dataset. More details of this dataset can be found in Appendix A. 
For object detection and semantic segmentation, we show results with all 3 baselines on the COCO 
% and Pascal VOC \texttt{trainval2017} 
datasets using Mask R-CNN, following \cite{chen2020improved}. We closely follow the common protocols listed in Detectron2 \cite{wu2019detectron2}.

\begin{table}[t]
% \begin{minipage}[b]{0.458\textwidth}
\centering
\setlength{\tabcolsep}{4pt}
    \begin{tabular}{lcccccc}
        \toprule
        % & \multicolumn{3}{c}{Detection} & \multicolumn{3}{c}{Segmentation} \\
        & AP$^{\rm b}$ & AP$^{\rm b}_{50}$ & AP$^{\rm b}_{75}$ & AP$^{\rm m}$ & AP$^{\rm m}_{50}$ & AP$^{\rm m}_{75}$\\
        \midrule
        % MoCo-v2 CC & 34.6 & 53.5 & 37.0 & 30.4 & 50.1 & 32.3\\
        % HCL/$\mathcal{L}_{\text{hyp}}$  CC &  36.1 & 55.2 & 37.9  & 31.5 & 52.0 & 33.8 \\
        % \rowcolor[HTML]{DFDFDF} 
        % HCL  CC  & \textbf{37.0} &  \textbf{56.1} & \textbf{39.8}  & \textbf{32.5} & \textbf{52.9}
        % &\textbf{34.6}\\ \hline
        \multicolumn{7}{l}{\textit{MoCo-v2 pre-trained on COCO:}}\\
        Baseline & 38.5 & 58.1 & 42.1 & 34.8 & 55.3 & 37.3\\
        Baseline + bbox &  39.7 & 60.1 & 43.4  & 36.0 & 57.3 & 38.8 \\
        \rowcolor[HTML]{DFDFDF} 
        HCL  & \textbf{40.6} &  \textbf{61.1} & \textbf{44.5}  & \textbf{37.0} & \textbf{58.3}
        &\textbf{39.7}\\ \hline
        \multicolumn{7}{l}{\textit{Dense-CL pre-trained on COCO:}}\\
        Baseline & 39.6 & 59.3 & 43.3 & 35.7 & 56.5 & 38.4\\
        Baseline + bbox & 41.3 & 	61.5 & 44.7 & 	37.5 & 59.5 & 40.4 \\
        \rowcolor[HTML]{DFDFDF} 
        HCL & \textbf{42.5} &  \textbf{62.5} & \textbf{45.8}  & \textbf{38.5} & \textbf{60.6}
        &\textbf{41.4}\\ \hline
        \multicolumn{7}{l}{\textit{ORL pre-trained on COCO:}}\\
        Baseline & 40.3 & 60.2 & 44.4 & 36.3 & 57.3	 & 38.9 \\
        \rowcolor[HTML]{DFDFDF} 
        HCL & \textbf{41.4} &  \textbf{61.4} & \textbf{45.5}  & \textbf{37.3} & \textbf{58.5}
        &\textbf{40.0}\\ \bottomrule
        \multicolumn{7}{l}{\textit{Dense-CL pre-trained on OpenImages:}}\\
        Baseline  & 38.2 & 58.9 & 42.6 & 34.8 & 55.3 & 37.8\\
        Baseline + bbox &  41.1 & 	61.5 & 44.4 & 	37.2 & 58.3 & 39.7 \\
        \rowcolor[HTML]{DFDFDF} 
        HCL & \textbf{42.1} &  \textbf{62.6} & \textbf{45.5}  & \textbf{38.3} & \textbf{59.4}
        &\textbf{40.6}\\ 
    \bottomrule
    \end{tabular}
    \captionof{table}{\textbf{Comparison with state-of-the-art methods.} This table shows object detection (columns 1-3) and semantic segmentation (columns 4-6) results on COCO using MoCo-v2, Dense-CL and ORL by pre-training on COCO and OpenImages using unsupervised object bounding boxes generated by the selective search. The first row in each sub-table shows the results using random crops on pre-training datasets. The second and third rows set HCL/$\mathcal{L}_{\text{hyp}}$ to 0, which  means we are pre-training baseline methods on just proposal boxes. Our model consistently improves both object detection and semantic segmentation tasks across multiple contrastive learning baselines by pre-training on both COCO (800 epochs) and the full OpenImages dataset (75 epochs, last 3 rows). 
    % Note that by only pre-training for 75 epochs we are getting better performance than is obtained by pre-training on ImageNet for 200 epochs using Dense-CL (40.3 mAP); this shows the importance of efficient pre-training on OpenImages.
    }
    \label{tab:voc_moco_200}
% \end{minipage}
\end{table}

\begin{table}
\begin{center}{
\setlength{\tabcolsep}{2.6pt}
    \begin{tabular}{lccccc}
        \toprule
        & Pre-train & Bbox & VOC & IN-100 & IN-1k \\
        \midrule
        MoCo-v2    & COCO & - & 64.79 & 64.84 &  51.17 \\
        MoCo-v2 + bbox   & COCO & SS & 73.13 & 73.84 & 54.21\\
        MoCo-v2 + bbox  & COCO & GT & 75.55 & 76.22 & 54.52\\
        \rowcolor[HTML]{DFDFDF} 
        HCL & COCO & SS & 74.19  & 75.16 &  55.03\\ 
        \rowcolor[HTML]{DFDFDF} 
        HCL & COCO & GT & \textbf{76.51} & \textbf{76.74} &  \textbf{55.63}\\
        \midrule
        MoCo-v2    & OpenImages & - & 69.95 & 72.80 & 54.12\\
        MoCo-v2 + bbox  & OpenImages & SS & 71.82 & 75.33 & 56.58\\
        MoCo-v2 + bbox  & OpenImages & GT & 73.79 & 77.36 & 57.57\\
        \rowcolor[HTML]{DFDFDF} 
        HCL & OpenImages & SS & 74.31 & 78.14 &  58.12\\
        \rowcolor[HTML]{DFDFDF} 
        HCL & OpenImages & GT & \textbf{75.40} & \textbf{79.08} &  \textbf{58.51}\\
        \bottomrule
        \end{tabular}
        \captionof{table}{\textbf{Classification results with linear evaluation.} The first row shows the results using random crops on pre-training datasets.
        % In the second and third row we set the HCL/$\mathcal{L}_{\text{hyp}}$ to 0, which  means we are pre-training MoCo-v2 on just proposal boxes. 
        In the last two rows we use our hyperbolic loss and we see improved performance by using both Ground Truth (GT) boxes and Selective Search (SS) boxes. HCL improves scene-level classification on the VOC dataset, and object-level classification on ImageNet-100 and ImageNet-1k datasets.}
    \label{tab:main_classification}}
    \end{center}
% \end{minipage}
\end{table}

\subsection{Main Results}
\topic{Object detection and semantic segmentation.} Table \ref{tab:voc_moco_200} reports the object detection and semantic segmentation results by pre-training on COCO and full OpenImages dataset (last 3 rows) by using selective search boxes. HCL shows consistent improvements over the baselines on COCO object detection and COCO semantic segmentation. Although Dense-CL and ORL improve the object-level downstream performance over MoCo-v2 through improved object representations or dense pixel representations, they still lack the direct modeling of scene images. We show that learning representations for scene images in hyperbolic space is beneficial to object-level downstream performance. Note that pre-training Dense-CL on ImageNet for 200 epochs gives 40.3 mAP~\cite{wang2021dense}, while pre-trainng on OpenImages for only 75 epochs with our method gives 42.1 mAP. 
This shows the importance of efficient pre-training on datasets like OpenImages.
% This section discusses our main results on the downstream image classification, object detection, and semantic segmentation tasks. 
% compare with the backbone model MoCo-v2~\cite{he2020momentum}. 

\topic{Image classification.}
As shown in Table \ref{tab:main_classification}, HCL improves image classification on both scene-level (VOC) and object-level (ImageNet) datasets. When pretraining on OpenImages, HCL improves ImageNet lineval accuracy by 0.94\% points and VOC lineval classification accuracy by 1.61 mAP. We observe similar improvements when pretraining on COCO. HCL improves accuracy whether we use ground truth object bounding boxes or boxes generated by selective search. In general, we observe a larger improvement of using HCL on OpenImages than COCO, which supports our hypothesis that HCL provides larger improvements on datasets with more objects per image.

% \vspace{-0.8in}

\subsection{Properties of Models Trained with HCL}
The visual representations learned by HCL have several useful properties. In this section, we evaluate the representation norm as an measure of the label uncertainty for image classification datasets, and evaluate the object-scene similarity in terms of out-of-context detection.

\subsubsection{Label Uncertainty Quantification}

% \centerline{\begin{minipage}[b]{0.98\textwidth}
% \centering
% \begin{minipage}[b]{0.38\textwidth}
%     \centering
\begin{table}
% \begin{minipage}[b]{0.458\textwidth}
\centering
    \includegraphics[trim=0 35 0 30, clip, width=0.95\textwidth]{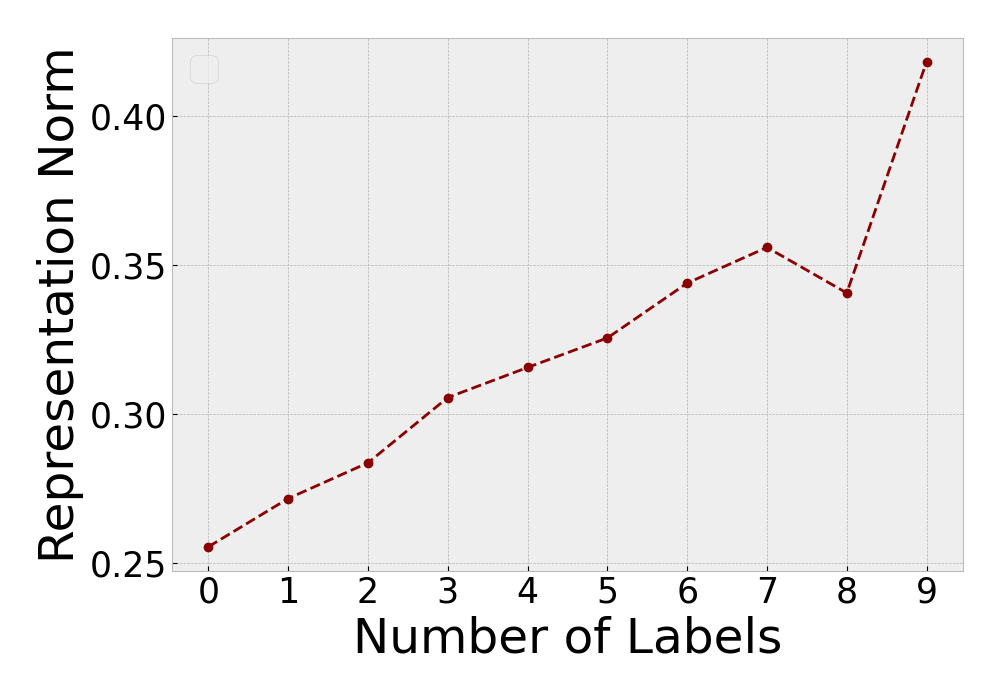}
    % \vspace{-0.4cm}
    \captionof{figure}{Average representation norms of images with different number of labels in ImageNet-ReaL.}
    \label{fig:in_real}
\end{table}

ImageNet~\cite{imagenet_cvpr09} is an image classification dataset consisting of object-centered images, each of which has a single label. As the performance on this dataset has gradually saturated, the original labels have been scrutinized more carefully~\cite{recht2019imagenet,tsipras2020imagenet,shankar2020evaluating,beyer2020imagenet,vasudevan2022does}. Prevailing labeling issues in the validation set have been recently identified, including labeling errors, multi-label images with only a single label provided, and so on. Although \cite{beyer2020imagenet} provides reassessed labels for the entire validation set, relabeling the entire training set may be infeasible. 

% \begin{minipage}[b]{0.58\textwidth}
% \centering
\begin{table}[t]
% \begin{minipage}[b]{0.458\textwidth}
\begin{center}
    \begin{tabular}{ll|ll}
    \toprule
    \multirow{2}{*}{Method} &  \multirow{2}{*}{Indicator} & \multicolumn{2}{c}{Datasets} \\
        &  & IN-Real & COCO \\
    \midrule
    MoCo     & Entropy         & 0.633 & 0.791 \\
    Supervised & Entropy         & 0.671 & 0.793 \\
    HCL      & Norm         & 0.655 & \textbf{0.839} \\
    Ensemble & Entropy+Norm       & \textbf{0.717} & 0.823 \\
    \bottomrule
    \end{tabular}
  \captionof{table}{NDCG scores of the image rankings based on the different indicators and models, and evaluated by the the number of labels per image.}
  \label{tab:ndcg}
\end{center}
\end{table}
\begin{figure*}[t]
    \centering
    % \vspace{-0.8cm}
    \includegraphics[width=\linewidth]{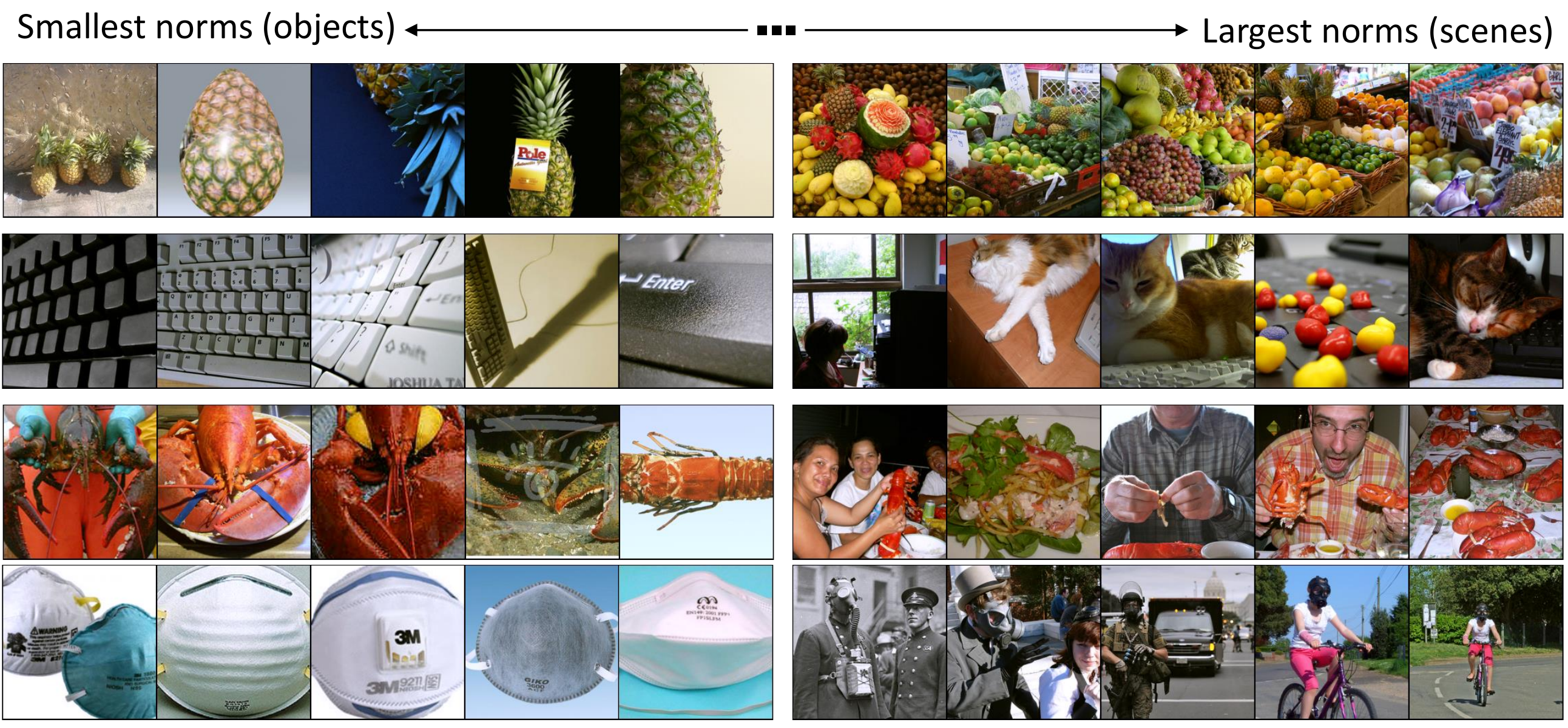}
    \caption{Images from ImageNet training set. The 5 images on the left have the smallest representation norms among all the images from the same class, and the 5 on the right have the largest norms.}
    \vspace{-0.2cm}
    \label{fig:imagenet_norm}
\end{figure*}

Our learned representations provide a potential automatic way to identify images with multiple labels from datasets like ImageNet. Specifically, we first show in Figure~\ref{fig:in_real} that there is a strong correlation between the representation norms and the number of labels per image according to the reassessed labels. For each class of the ImageNet training set, we use a pre-trained OpenImages model and rank the images according to their norms. The extreme images of some classes are shown in Figure~\ref{fig:imagenet_norm} and also in the Appendix. Images with smaller norms tend to capture a single object, while those with larger norms are likely to depict a scene. 
% In the keyboard class, we find that keyboards are often correlated with cats! 

To quantitatively evaluate this property, we report the NDCG metric on the ranked images as shown in Table~\ref{tab:ndcg}. NDCG assesses how often the scene images are ranked at the top. As a baseline, we rank the images based on the entropy of the class probability predicted by a classifier, which is a widely adopted indicator of label uncertainty~\cite{chen2019understanding,northcutt2021confident}. We use both MoCo-v2 and supervised ResNet-50 as the classifier. As shown in Table~\ref{tab:ndcg}, using norms with HCL achieves similar rank quality as using entropy with the supervised ResNet-50 on the ImageNet-ReaL dataset. In addition, when combining two ranks using simple ensemble methods such as Borda count, the score is further improved to $0.717$. This shows that the entropy and the norm provide complimentary signals regarding the existence of multiple labels. For example, the entropy indicator can be affected by the bias of the model and the norm indicator can be wrong on the images with multiple objects from the same class.

Compared to supervised indicators of label uncertainty, HCL has the additional advantage that it is dataset-agnostic and can be applied to new data without further training. To demonstrate this benefit, we report the same metric on the COCO validation, where we also have the number of labels for each image. Our method achieves much better NDCG scores than the supervised ResNet-50 as shown in Table~\ref{tab:ndcg}. This finding can be potentially useful to guide label reassessment, or provide an extra signal for model training.

\subsubsection{Out-of-Context Detection}

\begin{figure*}[h]
    \centering
    % \vspace{-0.8cm}
    \includegraphics[width=\linewidth]{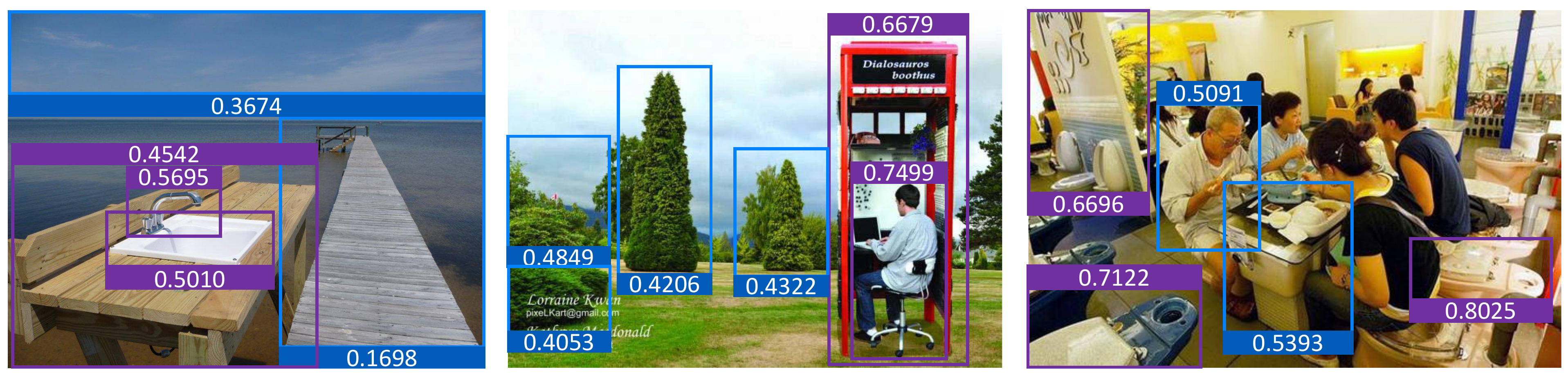}
    \caption{Out-of-context images from the SUN09 dataset. The bounding box of each object and its hyperbolic distance to the scene are shown. Regular objects are in blue and out-of-context objects are in purple. Note that the out-of-context objects tend to have large distances.}
    \label{fig:oot_detection}
\end{figure*}

Our hyperbolic loss $\mathcal{L}_{\text{hyp}}$ encourages the model to capture the similarity between the object and scene. We apply the resulting representations to detect out-of-context objects, which can be useful in designing data augmentation for object detection~\cite{dvornik2019importance}. We are especially interested in out-of-context images with conflicting backgrounds. To this end, we use the out-of-context images proposed in the SUN09 dataset~\cite{choi2010exploiting}. We first compute the representations of each object and entire scene image with that object masked out. We then calculate the hyperbolic distance between the representations mapped to the Poincaré ball. Some example images from this dataset as well as the distance of each contained object are shown in Figure~\ref{fig:oot_detection}. We find that the out-of-context objects generally have a large distance, i.e. smaller similarity, to the overall scene image. To quantify this finding, we compute the mAP of the object ranking on each image and obtain $0.61$ for HCL. As a comparison, the MoCo similarity gives mAP $=0.52$ and the random ranking gives mAP $=0.44$.

\section{Main Ablation Studies}
In this section, we report the results of several important ablation studies with respect to HCL. All the models are trained on the subset of the OpenImages dataset and linearly evaluated on the ImageNet-100 dataset. The top-1 accuracy is reported.

\begin{table*}[t]
\RawFloats
\small
\centering
\begin{minipage}[b]{0.4\textwidth}
\centering
    \begin{tabular}{llc}
    \toprule
    Distance  & Center & IN-100 Accuracy \\
    \midrule
    -          & -      & 77.36  \\
    \rowcolor[HTML]{DFDFDF} 
    Hyperbolic & Scene  & 79.08      \\
    Hyperbolic & Object & 76.96      \\
    Euclidean  & Scene  & 76.68     \\
    \bottomrule
    \end{tabular}
  \captionof{table}{Similarity measure and hierarchy center.}
    \label{tab:setup}
\end{minipage}%
\hfill
\begin{minipage}[b]{0.23\textwidth}
\centering
    \begin{tabular}{lc}
    \toprule
    $\lambda$ & IN-100 Accuracy \\
    \midrule
    0.01      & 77.70       \\
    \rowcolor[HTML]{DFDFDF} 
    0.1       & 79.08      \\
    0.2       & 78.64      \\
    0.5       & 0          \\
    \bottomrule
    \end{tabular}
  \captionof{table}{Losses trade-off.}
    \label{tab:lambda}
\end{minipage}
\hfill
\begin{minipage}[b]{0.36\textwidth}
\centering
    \begin{tabular}{llc}
    \toprule
    Optimizer & $\lambda$ & IN-100 Accuracy \\
    \midrule
    \rowcolor[HTML]{DFDFDF} 
    RSGD & 0.1 & 79.08  \\
    RSGD & 0.5 & 0      \\
    SGD  & 0.1 & 70.16  \\
    SGD  & 0.5 & 74.18  \\
    \bottomrule
    \end{tabular}
  \captionof{table}{RSGD versus SGD optimizers.}
    \label{tab:optim}
\end{minipage}
    \vspace{-0.5cm}
\end{table*}

\topic{Similarity measure and the center of the scene-object hierarchy.} We propose to use the negative hyperbolic distance as the similarity measure of the scene-object pairs. As an alternative, one can use cosine similarity on the hypersphere as the measure as in the original contrastive objective. However, this would attempt to maximize the similarity between a single object and multiple objects. It is likely that these objects belong to different classes, and hence this strategy impairs the quality of the representation. As shown in Table~\ref{tab:setup}, replacing the negative hyperbolic distance with the Euclidean similarity impairs downstream performance. The resulting model performs even worse than the baseline without loss function on the scene-object pairs, demonstrating the necessity of using hyperbolic distance. We also validate our choice of an object-centric hierarchy by comparing its performance with that of a scene-centric hierarchy~\cite{parikh2007hierarchical,parikh2009unsupervised} generated by sampling the negative pairs as objects and unpaired scenes. This scene-centric hierarchy leads to substantially lower accuracy (Table~\ref{tab:setup}). 

\topic{Trade-off between the Euclidean and hyperbolic losses.} We adopt the Euclidean loss to learn object-object similarity and the hyperbolic loss to learn object-scene similarity. A hyperparameter $\lambda$ controls the trade-off between them. As shown in Table~\ref{tab:lambda}, we find that a smaller $\lambda=0.01$ leads to marginal improvement. However, we also observe that larger $\lambda$s can lead to unstable and even stalled training. With careful inspection, we find that in the early stage of the training, the gradient provided by the hyperbolic loss can be inaccurate but strong, which pushes the representations to be close to the boundary. As a result, since Riemannian SGD divides gradients by the distance to the boundary, updates become small and training ceases to make progress.

\topic{Optimizer.} Given the observation above, we ask whether RSGD is necessary for practical usage. We replace the RSGD optimizer with SGD. To avoid numerical issues when the representations are too close to the boundary, we increase $\varepsilon$ from $1e^{-5}$ to $1e^{-1}$. This allows a larger $\lambda$ to be used as opposed to the RSGD. However, SGD always yields inferior performance compared to RSGD. %Therefore, it shows that the accurate gradient provided by RSGD is still necessary.

\vspace{-0.2cm}
\section{Related Work}
% \vspace{-0.1cm}
\topic{Representation Learning with Hyperbolic Space.}
% \vspace{-0.1cm}
Representations are typically learned in Euclidean space.
Hyperbolic space has been adopted for its expressiveness in modeling tree-like structures existing in various domains such as language~\cite{sala2018representation, nickel2017poincare, nickel2018learning}, graphs~\cite{balazevic2019multi,chami2020low,park2021unsupervised}, and vision~\cite{chen2020learning,suris2021learning}. The corresponding deep neural network modules have been designed to boost the progress of such applications~\cite{chami2019hyperbolic,ganea2018hyperbolicn,liu2019hyperbolic,shimizu2021hyperbolic}. The hierarchical structure presented in the datasets can arise from three factors that together motivate the use of hyperbolic space. The first factor is generality: the hypernym-hyponym property is a natural feature of words (e.g. WordNet~\cite{miller1990introduction}) and the hyperbolic space is extensively exploited to learn word and image embeddings that preserve that property~\cite{tifrea2018poincare, ganea2018hyperbolic, sala2018representation,liu2020hyperbolic,yan2021unsupervised,Long_2020_CVPR}. 
% Some image datasets also adopt the classes from WordNet for labeling, e.g. ImageNet~\cite{imagenet_cvpr09}, and consequently inherits the hierarchy in its labeling system. \citet{liu2020hyperbolic,yan2021unsupervised,Long_2020_CVPR} take advantage of hyperbolic space to capture such information in the visual embeddings. 
The second factor is uncertainty: Several studies have found that applying hyperbolic neural network modules to different tasks leads to a natural modeling of the uncertainty~\cite{ghadimiatigh2022hyperbolic,khrulkov2020hyperbolic,suris2021learning}. The third factor is compositionality of different basic elements to form a natural hierarchy. Motivated by these factors, previous work in computer vision has applied hierarchical representations learned in the hyperbolic space to various tasks such as image classification~\cite{khrulkov2020hyperbolic} or segmentation~\cite{weng2021unsupervised},  zero-/few-shot learning~\cite{liu2020hyperbolic}, action recognition~\cite{Long_2020_CVPR}, and video prediction~\cite{suris2021learning}. In this paper, we focus on learning the representations that capture the hierarchy between the objects and scenes with the goal of learning general-purpose image representations that can transfer to various downstream tasks.

\topic{Self-Supervised Learning on Scenes.} 
Self-Supervised Learning (SSL) has made great strides in closing the performance with supervised methods \cite{chen2020simple,chen2020improved,Ge2021RobustCL} when pretrained on the object-centric datasets like ImageNet. However, recent work has shown that SSL is limited on multi-object datasets like COCO \cite{selvaraju2017grad, wang2021dense, Mishra2020LearningVR} and OpenImages \cite{kuznetsova2020open}. Several papers have tried to address this issue by proposing different techniques. Dense-CL \cite{wang2021dense} operates on pre-average pool features and uses dense features on pixel level to show improved performance on dense tasks such as semantic segmentation. DetCon \cite{henaff2021efficient} uses unsupervised semantic segmentation masks to generate features for the corresponding objects in the two views. 
% CAST \cite{selvaraju2020casting} uses GradCAM \cite{selvaraju2017grad} to identify correspondences between objects across views and applies a contrastive loss on these features. 
PixContrast \cite{xie2021propagate} uses pixel-to-propagation consistency pretext task to build features for both dense downstream tasks and discriminative downstream tasks. Pixel-to-Pixel Contrast \cite{Wang_2021_ICCV} uses pixel-level contrastive learning to learn better features for semantic segmentation. Self-EMD \cite{liu2021selfemd} uses earth mover distance with BYOL \cite{grill2020bootstrap} for pretraining on the COCO dataset. ORL \cite{xie2021unsupervised} uses selective search to generate object proposals, then applies object-level contrastive loss to enforce object-level consistency. 
% ContraCAM \cite{NEURIPS2021_65d2ea03} mitigates scene bias by performing self-supervised object localization and applying a contrastive loss on the objects.
Below-par performance of SSL methods can be attributed to treating scenes and objects using similar techniques, which often results in similar representations. In our work, instead of treating scenes and objects similarly, we use a hyperbolic loss, which builds representation that disambiguates scenes and objects based on the norm of the embeddings. Our method not only separates scenes and objects, but also improves downstream tasks such as image classification.

% \dwj{Should you talk about scene work in general?  Could be fun to cite: Rosenfeld, Azriel, Robert A. Hummel, and Steven W. Zucker. "Scene labeling by relaxation operations." IEEE Transactions on Systems, Man, and Cybernetics 6 (1976): 420-433.}

\section{Conclusion}
% \topic{Conclusion} 
We present HCL, a contrastive learning framework that learns visual representation for both objects and scenes in the same representation space. The major novelty of our method is a hyperbolic contrastive objective built on an object-centric scene hierarchy. We show the effectiveness of HCL on several benchmarks including image classification, object detection, and semantic segmentation. We also demonstrate useful properties of the representations under several zero-shot settings, from detecting out-of-context objects to quantifying the label uncertainty in the datasets like ImageNet. More generally, we hope this paper will encourage future work towards building a more holistic visual representation space, and draw attention to the power of non-Euclidean representation learning.

\section{Acknowledgements}
Songwei Ge, Shlok Mishra, David Jacobs were supported in part by the National Science Foundation under grant no. IIS-1910132 and IIS-2213335.

% \topic{Limitations} We have shown that our model improves classification performance on the ImageNet dataset, but gains on more fine-grained classification tasks are smaller, as shown in Appendix B.2. We conjecture that our model primarily improves object representations by modeling context information, whereas most of these fine-grained datasets have similar contextual information. In addition, although we demonstrate the importance of Riemannian optimization, its underlying mechanism in the visual representation learning remains unclear. We conduct more experiments on training hyperbolic linear classifiers in Appendix C.1. However, further efforts are needed to fully unleash the potential of non-Euclidean representation learning.



% {\small
% \bibliographystyle{ieee_fullname}
% \bibliography{iclr2023_conference}
% }
%%%%%%%%%%%%%%%%%%%%%%%%%%%%%%%%%%%%%%%%%%%%%%%%%%%%%%%%%%%%

% \clearpage
\appendix
\section{Experiment setups}

In this section, we provide additional details of our experiments.

\paragraph{Unsupervised object proposals.} When pretraining on uncurated datasets, acquiring ground truth object bounding boxes using human annotations can be expensive. However, automatically generating unsupervised region proposals is well-studied. We use Selective Search as the unsupervised proposal generation method. Following ORL \cite{xie2021unsupervised} we first generate the proposals using selective search. Then we filter the proposals with 96 pixels as the minimal scale, maximum IOU of 0.5 and  aspect ratio between 1/3 to 3. For every image we generate maximum of 100 proposals and randomly select any image as the object image.

\paragraph{OpenImages dataset.} We use the full OpenImages dataset which have bounding box annotations ($\sim$ 1.9 million images). We also use a subset proposed in \cite{Mishra2021ObjectAwareCF}. This is a subset created from the OpenImages dataset where each image has at least 2 classes present and each class has at least 900 instances. This subset is a balanced subset of OpenImages with an average of 12 object present in an image, making it a good proxy for real-world multi-object images. 

% \paragraph{INPMix dataset.} We sample a subset of the Place-205~\cite{zhou2014learning} dataset to test the discriminative capacity of the representations on the scene images. Specifically, we randomly sample $1,300$ training images from each of the $205$ classes, and then combine them with the ImageNet-100~\cite{tian2019contrastive} to form a dataset of $305$ classes in total. We provide the code to reproduce the dataset in the supplementary material.

\paragraph{Object and Scene image augmentations.} We find that small objects are always detrimental to performance. Therefore, when sampling object bounding boxes, we drop bounding boxes with size \texttt{width} $\times$ \texttt{height} $\le 56 \times 56$. Further, when sampling objects for the Euclidean branch, if the size of a bounding box \texttt{width} $\times$ \texttt{height} $\le 256 \times 256$, we slightly expand it to either $256 \times 256$ or the maximal size allowed by the original image size. We also apply a small jittering to the  \texttt{width} and \texttt{height} to include different contexts around the objects. Next, we apply random cropping and resizing with the same scale $(0.2, 1.)$ as in MoCo~\cite{he2020momentum}. When sampling objects for the hyperbolic branch, we do not apply jittering and random cropping, but only filter the small boxes and resize to $\le 224 \times 224$. To crop the scene images, we sample another $1$ to $5$ bounding boxes and merge with the selected object bounding box.  

\paragraph{Model details of pre-training.} For the optimizer setups and augmentation recipes, we follow the standard protocol described in MoCo-v2~\cite{chen2020improved}. We find that a base learning rate of 0.3 works better when pre-training on COCO and OpenImage datasets as compared to 0.03. We adopt the linear learning rate scaling receipt that $lr=0.3 \times \text{BatchSize}/256$~\cite{goyal2017accurate} and batch size of $128$ by default on $4$ NVIDIA p6000 gpus. To ensure fair comparison, we also pre-train the baselines with a learning rate of 0.3. We train our models on COCO and the subset of OpenImage datasets for $200$ epochs and full OpenImage dataset for $75$ epochs. We also note that calculating hyperbolic loss itself takes nearly the same time as a normal contrastive loss. The only overhead in pre-training is one additional forward pass to get scene representations. In our setting, MoCo takes 0.616 sec/iter while HCL takes 0.757 sec/iter. For the hyperparameters of our hyperbolic objective, we use $r=4.5$, $\lambda=0.1$, and $\varepsilon=1e^{-5}$ as our default setting.

\section{Additional experimental results}

\subsection{Robustness under Corruption.}

We calculate the mCE error as in Hendrycks et al. \cite{hendrycks2019robustness}. We compare our HCL model trained on OpenImages and lineval on ImageNet dataset with the baseline model without using HCL loss. We see an improvement of 1.9 mCE over the baseline model, demonstrating that our HCL model learns more robust representations as compared to the vanilla MoCo. 
% More detailed results are present in the supplementary materials.

\subsection{Fine-grained class classification}
\begin{table}[h]
    \centering
    \begin{tabular}{lccc}
        \toprule
        Method  & Cars~\cite{KrauseStarkDengFei-Fei_3DRR2013} & DTD~\cite{cimpoi14describing} &  Food~\cite{bossard14}\\
        \midrule
        HCL/$\mathcal{L}_{\text{hyp}}$    &   31.92 &	68.46 &	58.66  \\
        \rowcolor[HTML]{DFDFDF} 
        HCL   & 32.02 &	68.19 &	58.79\\
    \bottomrule
    \end{tabular}
    \captionof{table}{Fine grained classification results.}
    \label{tab:fine_grained}
\end{table}

In Table \ref{tab:fine_grained} we show results on fine-grained classification datasets. We can see that on fine-grained classification our model provides little performance improvement. This could be due to the fact that all classes in these datasets have very similar scene contexts, and hence the hyperbolic objective does not help very much.

\subsection{More ImageNet Examples}

% \vspace{-0.3cm}
\begin{figure*}[h]
    \centering
    \includegraphics[width=0.95\linewidth]{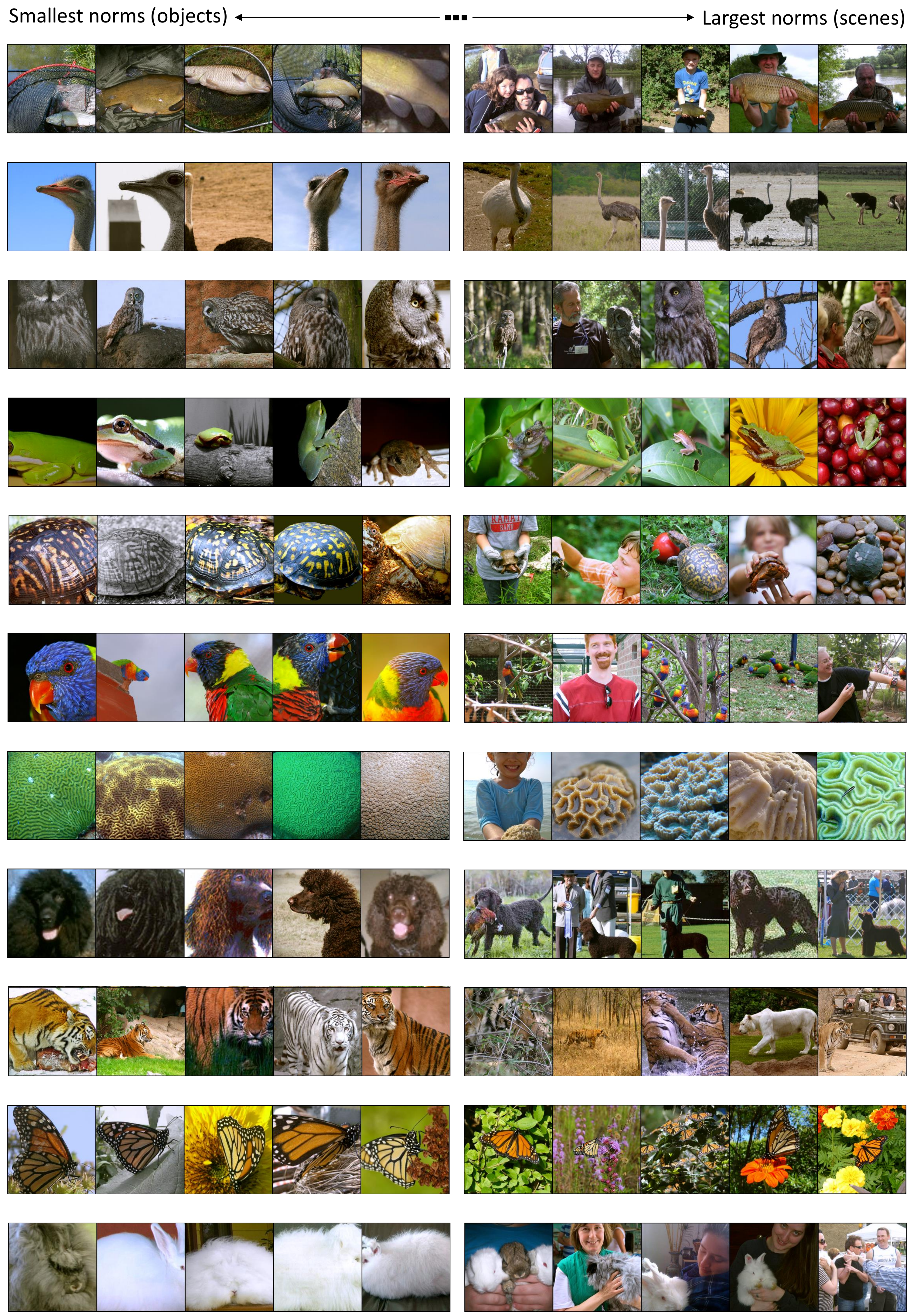}
    \caption{More images from ImageNet training set sorted by their representation norms.}
\end{figure*}
\clearpage
\begin{figure*}
    \centering
    \includegraphics[width=0.95\linewidth]{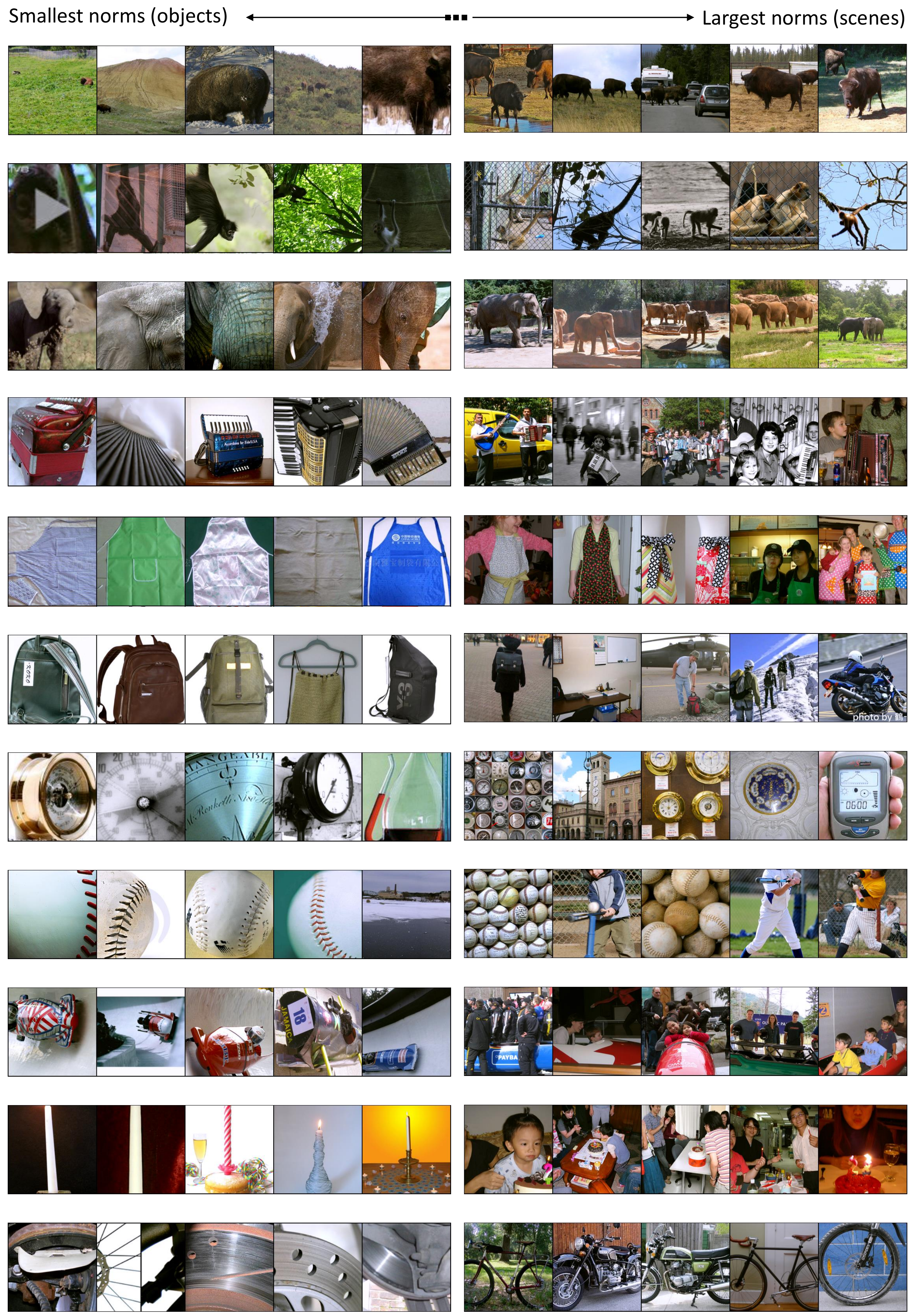}
    \caption{More images from ImageNet training set sorted by their representation norms.}
\end{figure*}
\clearpage

\section{Additional ablation studies}
In this section, we provide more ablation experiments on hyperbolic linear evaluation, model architecture, and the radius of Poincare ball. All the models are trained on the OpenImages dataset and evaluated on the ImageNet-100 (IN-100) or ImageNet-1k (IN-1k) the top-1 accuracy reported.

\paragraph{Radius of the Poincare ball.}
In Table \ref{tab:poincare_radius} we show results by varying the radius of Poincaré ball. The hyperbolic objective improves the performance over all the tested radius. We find that a too small radius may lead to a smaller improvement due to the stronger regularization. 

\paragraph{Configuration of the encoder head.}
In our experiments, the Euclidean and hyperbolic branches share the weights in both the backbone and the head of the encoders. We also try using a separate head for the hyperbolic branch. As shown in Table~\ref{tab:predictor}, this leads to a more stable training when larger learning rate is applied. However, we did not see any improvements brought by this modification.

\paragraph{Hyperbolic linear evaluation.}
Apart from the common linear evaluation in the Euclidean space, we show the hyperbolic linear evaluation results with different optimizers and learning rates in Table \ref{tab:hyperbolic_lineval}. The idea is to test if the representations are more linearly separable in the hyperbolic space. We follow the same setting of hyperbolic softmax regression~\cite{ganea2018hyperbolicn} and train a single hyperbolic linear layer. However, we find the optimization with SGD can easily cause overflow. By contrast, Adam is much more stable with appropriate learning rates.

\begin{minipage}[b]{0.25\textwidth}
\centering
    \begin{tabular}{lccc}
        \toprule
        $c$ & IN-1k Acc. \\
        \midrule
        1 &	58.08 \\
        0.5	& 58.31 \\
        0.1 &	58.29 \\
        \rowcolor[HTML]{DFDFDF} 
        0.05 &	58.51 \\
        0.01 &	58.49 \\
    \bottomrule
    \end{tabular}
    \captionof{table}{Results by varying the radius $r$ of Poincaré ball. $c=\frac{1}{r^2}.$}
    \label{tab:poincare_radius}
\end{minipage}
\hfill
\begin{minipage}[b]{0.33\textwidth}
\centering
    \begin{tabular}{lccc}
        \toprule
        \multicolumn{1}{l}{Head} & $\lambda$ & IN-100 Acc.\\
        \midrule
        \multicolumn{1}{l}{N/A}        & 0         & 77.36  \\
        \rowcolor[HTML]{DFDFDF} 
        \multirow{2}{*}{shared}        & 0.1       & 79.08  \\
                                       & 0.5       & 0      \\
        \multirow{2}{*}{splitted}      & 0.1       & 77.88  \\
                                       & 0.5       & 77.58  \\
    \bottomrule
    \end{tabular}
    \captionof{table}{Different configurations of head in the the Euclidean and hyperbolic branches.}
    \label{tab:predictor}
\end{minipage}
\hfill
\begin{minipage}[b]{0.35\textwidth}
\centering
    \begin{tabular}{llll}
    \toprule
    \multicolumn{2}{c}{SGD} & \multicolumn{2}{c}{Adam} \\
    lr & IN-100 & lr & IN-100 \\
    \midrule
    0.1        & 63.82       & 0.001        & 67.64      \\
    0.2        & 64.22       & 0.0005       & 70.32      \\
    0.3        & 1           & 0.0001       & 72.58      \\
    0.4        & 1           & 0.00005      & 70.5       \\
    \bottomrule
    \end{tabular}
    \captionof{table}{Results of hyperbolic linear evaluation with different optimizers and learning rates. }
    \label{tab:hyperbolic_lineval}
\end{minipage}

{\small
\bibliographystyle{ieee_fullname}
\bibliography{iclr2023_conference}

\begin{thebibliography}{10}\itemsep=-1pt

\bibitem{bai2022point}
Yutong Bai, Xinlei Chen, Alexander Kirillov, Alan Yuille, and Alexander~C Berg.
\newblock Point-level region contrast for object detection pre-training.
\newblock {\em CVPR}, 2022.

\bibitem{balazevic2019multi}
Ivana Balazevic, Carl Allen, and Timothy Hospedales.
\newblock Multi-relational poincar{\'e} graph embeddings.
\newblock {\em NeurIPS}, 2019.

\bibitem{beyer2020imagenet}
Lucas Beyer, Olivier~J. Hénaff, Alexander Kolesnikov, Xiaohua Zhai, and Aäron
  van~den Oord.
\newblock Are we done with imagenet?, 2020.

\bibitem{bonnabel2013stochastic}
Silvere Bonnabel.
\newblock Stochastic gradient descent on riemannian manifolds.
\newblock {\em IEEE Transactions on Automatic Control}, 58(9):2217--2229, 2013.

\bibitem{bossard14}
Lukas Bossard, Matthieu Guillaumin, and Luc Van~Gool.
\newblock Food-101 -- mining discriminative components with random forests.
\newblock In {\em ECCV}, 2014.

\bibitem{cannon1997hyperbolic}
James~W Cannon, William~J Floyd, Richard Kenyon, Walter~R Parry, et~al.
\newblock Hyperbolic geometry.
\newblock {\em Flavors of geometry}, 31(59-115):2, 1997.

\bibitem{caron2020unsupervised}
Mathilde Caron, Ishan Misra, Julien Mairal, Priya Goyal, Piotr Bojanowski, and
  Armand Joulin.
\newblock Unsupervised learning of visual features by contrasting cluster
  assignments.
\newblock {\em NeurIPS}, 2020.

\bibitem{caron2021emerging}
Mathilde Caron, Hugo Touvron, Ishan Misra, Herv{\'e} J{\'e}gou, Julien Mairal,
  Piotr Bojanowski, and Armand Joulin.
\newblock Emerging properties in self-supervised vision transformers.
\newblock In {\em ICCV}, 2021.

\bibitem{chami2020low}
Ines Chami, Adva Wolf, Da-Cheng Juan, Frederic Sala, Sujith Ravi, and
  Christopher R{\'e}.
\newblock Low-dimensional hyperbolic knowledge graph embeddings.
\newblock In {\em ACL}, 2020.

\bibitem{chami2019hyperbolic}
Ines Chami, Zhitao Ying, Christopher R{\'e}, and Jure Leskovec.
\newblock Hyperbolic graph convolutional neural networks.
\newblock {\em NeurIPS}, 2019.

\bibitem{chen2020learning}
Jiaxin Chen, Jie Qin, Yuming Shen, Li Liu, Fan Zhu, and Ling Shao.
\newblock Learning attentive and hierarchical representations for 3d shape
  recognition.
\newblock In {\em ECCV}, 2020.

\bibitem{chen2019understanding}
Pengfei Chen, Ben~Ben Liao, Guangyong Chen, and Shengyu Zhang.
\newblock Understanding and utilizing deep neural networks trained with noisy
  labels.
\newblock In {\em ICML}, 2019.

\bibitem{chen2020simple}
Ting Chen, Simon Kornblith, Mohammad Norouzi, and Geoffrey Hinton.
\newblock A simple framework for contrastive learning of visual
  representations.
\newblock In {\em ICML}, 2020.

\bibitem{chen2020improved}
Xinlei Chen, Haoqi Fan, Ross Girshick, and Kaiming He.
\newblock Improved baselines with momentum contrastive learning.
\newblock {\em arXiv preprint arXiv:2003.04297}, 2020.

\bibitem{choi2010exploiting}
Myung~Jin Choi, Joseph~J Lim, Antonio Torralba, and Alan~S Willsky.
\newblock Exploiting hierarchical context on a large database of object
  categories.
\newblock In {\em CVPR}, 2010.

\bibitem{cimpoi14describing}
M. Cimpoi, S. Maji, I. Kokkinos, S. Mohamed, , and A. Vedaldi.
\newblock Describing textures in the wild.
\newblock In {\em CVPR}, 2014.

\bibitem{imagenet_cvpr09}
Jia Deng, Wei Dong, Richard Socher, Li-Jia Li, Kai Li, and Li Fei-Fei.
\newblock Imagenet: A large-scale hierarchical image database.
\newblock In {\em CVPR}, 2009.

\bibitem{do1992riemannian}
Manfredo~Perdigao Do~Carmo and J Flaherty~Francis.
\newblock {\em Riemannian geometry}, volume~6.
\newblock Springer, 1992.

\bibitem{dvornik2019importance}
Nikita Dvornik, Julien Mairal, and Cordelia Schmid.
\newblock On the importance of visual context for data augmentation in scene
  understanding.
\newblock {\em PAMI}, 43(6):2014--2028, 2019.

\bibitem{everingham2010pascal}
Mark Everingham, Luc Van~Gool, Christopher~KI Williams, John Winn, and Andrew
  Zisserman.
\newblock The pascal visual object classes (voc) challenge.
\newblock {\em IJCV}, 88(2):303--338, 2010.

\bibitem{galleguillos2008object}
Carolina Galleguillos, Andrew Rabinovich, and Serge Belongie.
\newblock Object categorization using co-occurrence, location and appearance.
\newblock In {\em CVPR}, 2008.

\bibitem{ganea2018hyperbolic}
Octavian Ganea, Gary B{\'e}cigneul, and Thomas Hofmann.
\newblock Hyperbolic entailment cones for learning hierarchical embeddings.
\newblock In {\em ICML}, 2018.

\bibitem{ganea2018hyperbolicn}
Octavian Ganea, Gary B{\'e}cigneul, and Thomas Hofmann.
\newblock Hyperbolic neural networks.
\newblock {\em NeurIPS}, 2018.

\bibitem{Ge2021RobustCL}
Songwei Ge, Shlok~Kumar Mishra, Haohan Wang, Chun-Liang Li, and David Jacobs.
\newblock Robust contrastive learning using negative samples with diminished
  semantics.
\newblock In {\em NeurIPS}, 2021.

\bibitem{ghadimiatigh2022hyperbolic}
Mina GhadimiAtigh, Julian Schoep, Erman Acar, Nanne van Noord, and Pascal
  Mettes.
\newblock Hyperbolic image segmentation.
\newblock {\em arXiv preprint arXiv:2203.05898}, 2022.

\bibitem{goyal2017accurate}
Priya Goyal, Piotr Doll{\'a}r, Ross Girshick, Pieter Noordhuis, Lukasz
  Wesolowski, Aapo Kyrola, Andrew Tulloch, Yangqing Jia, and Kaiming He.
\newblock Accurate, large minibatch sgd: Training imagenet in 1 hour.
\newblock {\em arXiv preprint arXiv:1706.02677}, 2017.

\bibitem{grill2020bootstrap}
Jean-Bastien Grill, Florian Strub, Florent Altch\'{e}, Corentin Tallec, Pierre
  Richemond, Elena Buchatskaya, Carl Doersch, Bernardo Avila~Pires, Zhaohan
  Guo, Mohammad Gheshlaghi~Azar, Bilal Piot, koray kavukcuoglu, Remi Munos, and
  Michal Valko.
\newblock Bootstrap your own latent - a new approach to self-supervised
  learning.
\newblock In {\em NeurIPS}, 2020.

\bibitem{gromov1987hyperbolic}
Mikhael Gromov.
\newblock Hyperbolic groups.
\newblock In {\em Essays in group theory}, pages 75--263. Springer, 1987.

\bibitem{He_2022_CVPR}
Kaiming He, Xinlei Chen, Saining Xie, Yanghao Li, Piotr Doll\'ar, and Ross
  Girshick.
\newblock Masked autoencoders are scalable vision learners.
\newblock In {\em Proceedings of the IEEE/CVF Conference on Computer Vision and
  Pattern Recognition (CVPR)}, pages 16000--16009, June 2022.

\bibitem{he2020momentum}
Kaiming He, Haoqi Fan, Yuxin Wu, Saining Xie, and Ross Girshick.
\newblock Momentum contrast for unsupervised visual representation learning.
\newblock In {\em CVPR}, 2020.

\bibitem{henaff2021efficient}
Olivier~J H{\'e}naff, Skanda Koppula, Jean-Baptiste Alayrac, Aaron van~den
  Oord, Oriol Vinyals, and Jo{\~a}o Carreira.
\newblock Efficient visual pretraining with contrastive detection.
\newblock In {\em ICCV}, pages 10086--10096, 2021.

\bibitem{hendrycks2019robustness}
Dan Hendrycks and Thomas Dietterich.
\newblock Benchmarking neural network robustness to common corruptions and
  perturbations.
\newblock {\em ICLR}, 2019.

\bibitem{hinton2021represent}
Geoffrey Hinton.
\newblock How to represent part-whole hierarchies in a neural network.
\newblock {\em arXiv preprint arXiv:2102.12627}, 2021.

\bibitem{johnson2015image}
Justin Johnson, Ranjay Krishna, Michael Stark, Li-Jia Li, David Shamma, Michael
  Bernstein, and Li Fei-Fei.
\newblock Image retrieval using scene graphs.
\newblock In {\em CVPR}, 2015.

\bibitem{khrulkov2020hyperbolic}
Valentin Khrulkov, Leyla Mirvakhabova, Evgeniya Ustinova, Ivan Oseledets, and
  Victor Lempitsky.
\newblock Hyperbolic image embeddings.
\newblock In {\em CVPR}, 2020.

\bibitem{KrauseStarkDengFei-Fei_3DRR2013}
Jonathan Krause, Michael Stark, Jia Deng, and Li Fei-Fei.
\newblock 3d object representations for fine-grained categorization.
\newblock In {\em 3DRR}, 2013.

\bibitem{krishna2017visual}
Ranjay Krishna, Yuke Zhu, Oliver Groth, Justin Johnson, Kenji Hata, Joshua
  Kravitz, Stephanie Chen, Yannis Kalantidis, Li-Jia Li, David~A Shamma, et~al.
\newblock Visual genome: Connecting language and vision using crowdsourced
  dense image annotations.
\newblock {\em IJCV}, 2017.

\bibitem{kuznetsova2020open}
Alina Kuznetsova, Hassan Rom, Neil Alldrin, Jasper Uijlings, Ivan Krasin, Jordi
  Pont-Tuset, Shahab Kamali, Stefan Popov, Matteo Malloci, Alexander
  Kolesnikov, et~al.
\newblock The open images dataset v4.
\newblock {\em IJCV}, 2020.

\bibitem{lee2018introduction}
John~M Lee.
\newblock {\em Introduction to Riemannian manifolds}.
\newblock Springer, 2018.

\bibitem{Lin2014MicrosoftCC}
Tsung-Yi Lin, M. Maire, Serge~J. Belongie, James Hays, P. Perona, D. Ramanan,
  Piotr Doll{\'a}r, and C.~L. Zitnick.
\newblock Microsoft coco: Common objects in context.
\newblock In {\em ECCV}, 2014.

\bibitem{linial1995geometry}
Nathan Linial, Eran London, and Yuri Rabinovich.
\newblock The geometry of graphs and some of its algorithmic applications.
\newblock {\em Combinatorica}, 15(2):215--245, 1995.

\bibitem{liu2019hyperbolic}
Qi Liu, Maximilian Nickel, and Douwe Kiela.
\newblock Hyperbolic graph neural networks.
\newblock {\em NeurIPS}, 2019.

\bibitem{liu2020hyperbolic}
Shaoteng Liu, Jingjing Chen, Liangming Pan, Chong-Wah Ngo, Tat-Seng Chua, and
  Yu-Gang Jiang.
\newblock Hyperbolic visual embedding learning for zero-shot recognition.
\newblock In {\em CVPR}, 2020.

\bibitem{liu2021selfemd}
Songtao Liu, Zeming Li, and Jian Sun.
\newblock Self-emd: Self-supervised object detection without imagenet, 2021.

\bibitem{Long_2020_CVPR}
Teng Long, Pascal Mettes, Heng~Tao Shen, and Cees G.~M. Snoek.
\newblock Searching for actions on the hyperbole.
\newblock In {\em CVPR}, 2020.

\bibitem{mensink2014costa}
Thomas Mensink, Efstratios Gavves, and Cees~GM Snoek.
\newblock Costa: Co-occurrence statistics for zero-shot classification.
\newblock In {\em CVPR}, 2014.

\bibitem{miller1990introduction}
George~A Miller, Richard Beckwith, Christiane Fellbaum, Derek Gross, and
  Katherine~J Miller.
\newblock Introduction to wordnet: An on-line lexical database.
\newblock {\em International journal of lexicography}, 3(4):235--244, 1990.

\bibitem{Mishra2020LearningVR}
Shlok~Kumar Mishra, Anshul~B. Shah, Ankan Bansal, Jonghyun Choi, Abhinav
  Shrivastava, Abhishek Sharma, and David Jacobs.
\newblock Learning visual representations for transfer learning by suppressing
  texture.
\newblock {\em ArXiv}, abs/2011.01901, 2020.

\bibitem{Mishra2021ObjectAwareCF}
Shlok~Kumar Mishra, Anshul~B. Shah, Ankan Bansal, Abhyuday~N. Jagannatha,
  Abhishek Sharma, David Jacobs, and Dilip Krishnan.
\newblock Object-aware cropping for self-supervised learning.
\newblock {\em ArXiv}, abs/2112.00319, 2021.

\bibitem{nickel2017poincare}
Maximillian Nickel and Douwe Kiela.
\newblock Poincar{\'e} embeddings for learning hierarchical representations.
\newblock {\em NeurIPS}, 2017.

\bibitem{nickel2018learning}
Maximillian Nickel and Douwe Kiela.
\newblock Learning continuous hierarchies in the lorentz model of hyperbolic
  geometry.
\newblock In {\em ICML}, 2018.

\bibitem{northcutt2021confident}
Curtis Northcutt, Lu Jiang, and Isaac Chuang.
\newblock Confident learning: Estimating uncertainty in dataset labels.
\newblock {\em Journal of Artificial Intelligence Research}, 70:1373--1411,
  2021.

\bibitem{parikh2007hierarchical}
Devi Parikh and Tsuhan Chen.
\newblock Hierarchical semantics of objects (hsos).
\newblock In {\em ICCV}, 2007.

\bibitem{parikh2009unsupervised}
Devi Parikh, C~Lawrence Zitnick, and Tsuhan Chen.
\newblock Unsupervised learning of hierarchical spatial structures in images.
\newblock In {\em CVPR}, 2009.

\bibitem{park2021unsupervised}
Jiwoong Park, Junho Cho, Hyung~Jin Chang, and Jin~Young Choi.
\newblock Unsupervised hyperbolic representation learning via message passing
  auto-encoders.
\newblock In {\em CVPR}, 2021.

\bibitem{radford2021learning}
Alec Radford, Jong~Wook Kim, Chris Hallacy, Aditya Ramesh, Gabriel Goh,
  Sandhini Agarwal, Girish Sastry, Amanda Askell, Pamela Mishkin, Jack Clark,
  et~al.
\newblock Learning transferable visual models from natural language
  supervision.
\newblock In {\em ICML}, 2021.

\bibitem{recht2019imagenet}
Benjamin Recht, Rebecca Roelofs, Ludwig Schmidt, and Vaishaal Shankar.
\newblock Do imagenet classifiers generalize to imagenet?
\newblock In {\em ICML}, 2019.

\bibitem{sala2018representation}
Frederic Sala, Chris De~Sa, Albert Gu, and Christopher R{\'e}.
\newblock Representation tradeoffs for hyperbolic embeddings.
\newblock In {\em ICML}, 2018.

\bibitem{selvaraju2017grad}
Ramprasaath~R Selvaraju, Michael Cogswell, Abhishek Das, Ramakrishna Vedantam,
  Devi Parikh, and Dhruv Batra.
\newblock Grad-cam: Visual explanations from deep networks via gradient-based
  localization.
\newblock In {\em ICCV}, 2017.

\bibitem{shankar2020evaluating}
Vaishaal Shankar, Rebecca Roelofs, Horia Mania, Alex Fang, Benjamin Recht, and
  Ludwig Schmidt.
\newblock Evaluating machine accuracy on imagenet.
\newblock In {\em ICML}, 2020.

\bibitem{shimizu2021hyperbolic}
Ryohei Shimizu, YUSUKE Mukuta, and Tatsuya Harada.
\newblock Hyperbolic neural networks++.
\newblock In {\em ICLR}, 2021.

\bibitem{suris2021learning}
D{\'\i}dac Sur{\'\i}s, Ruoshi Liu, and Carl Vondrick.
\newblock Learning the predictability of the future.
\newblock In {\em CVPR}, 2021.

\bibitem{tian2019contrastive}
Yonglong Tian, Dilip Krishnan, and Phillip Isola.
\newblock Contrastive multiview coding.
\newblock In {\em ECCV}, 2020.

\bibitem{tifrea2018poincare}
Alexandru Tifrea, Gary B{\'e}cigneul, and Octavian-Eugen Ganea.
\newblock Poincar{\'e} glove: Hyperbolic word embeddings.
\newblock In {\em ICLR}. OpenReview, 2018.

\bibitem{tsipras2020imagenet}
Dimitris Tsipras, Shibani Santurkar, Logan Engstrom, Andrew Ilyas, and
  Aleksander Madry.
\newblock From imagenet to image classification: Contextualizing progress on
  benchmarks.
\newblock In {\em ICML}, 2020.

\bibitem{uijlings2013selective}
Jasper~RR Uijlings, Koen~EA Van De~Sande, Theo Gevers, and Arnold~WM Smeulders.
\newblock Selective search for object recognition.
\newblock {\em IJCV}, 104(2):154--171, 2013.

\bibitem{vasudevan2022does}
Vijay Vasudevan, Benjamin Caine, Raphael Gontijo-Lopes, Sara Fridovich-Keil,
  and Rebecca Roelofs.
\newblock When does dough become a bagel? analyzing the remaining mistakes on
  imagenet.
\newblock {\em arXiv preprint arXiv:2205.04596}, 2022.

\bibitem{Wang_2021_ICCV}
Wenguan Wang, Tianfei Zhou, Fisher Yu, Jifeng Dai, Ender Konukoglu, and Luc
  Van~Gool.
\newblock Exploring cross-image pixel contrast for semantic segmentation.
\newblock In {\em ICCV}, 2021.

\bibitem{wang2021dense}
Xinlong Wang, Rufeng Zhang, Chunhua Shen, Tao Kong, and Lei Li.
\newblock Dense contrastive learning for self-supervised visual pre-training.
\newblock In {\em CVPR}, 2021.

\bibitem{weng2021unsupervised}
Zhenzhen Weng, Mehmet~Giray Ogut, Shai Limonchik, and Serena Yeung.
\newblock Unsupervised discovery of the long-tail in instance segmentation
  using hierarchical self-supervision.
\newblock In {\em CVPR}, 2021.

\bibitem{wu2019detectron2}
Yuxin Wu, Alexander Kirillov, Francisco Massa, Wan-Yen Lo, and Ross Girshick.
\newblock Detectron2.
\newblock \url{https://github.com/facebookresearch/detectron2}, 2019.

\bibitem{wu2018unsupervised}
Zhirong Wu, Yuanjun Xiong, Stella~X Yu, and Dahua Lin.
\newblock Unsupervised feature learning via non-parametric instance
  discrimination.
\newblock In {\em CVPR}, 2018.

\bibitem{xie2021unsupervised}
Jiahao Xie, Xiaohang Zhan, Ziwei Liu, Yew~Soon Ong, and Chen~Change Loy.
\newblock Unsupervised object-level representation learning from scene images.
\newblock In {\em NeurIPS}, 2021.

\bibitem{xie2021propagate}
Zhenda Xie, Yutong Lin, Zheng Zhang, Yue Cao, Stephen Lin, and Han Hu.
\newblock Propagate yourself: Exploring pixel-level consistency for
  unsupervised visual representation learning.
\newblock In {\em CVPR}, 2021.

\bibitem{yan2021unsupervised}
Jiexi Yan, Lei Luo, Cheng Deng, and Heng Huang.
\newblock Unsupervised hyperbolic metric learning.
\newblock In {\em CVPR}, 2021.

\end{thebibliography}
}

\end{document}